\documentclass[conference]{IEEEtran}
\IEEEoverridecommandlockouts
\usepackage{cite}
\usepackage{amsmath,amssymb,amsfonts}
\usepackage{listings}
\usepackage{algorithmic}
\usepackage{graphicx}
\usepackage{textcomp}
\usepackage{xcolor}
\usepackage{url}
\usepackage{verbatim}
\usepackage{hyperref}
\usepackage{tikz}
\def\BibTeX{{\rm B\kern-.05em{\sc i\kern-.025em b}\kern-.08em
    T\kern-.1667em\lower.7ex\hbox{E}\kern-.125emX}}
    
    \definecolor{lightGray}{gray}{0.9}
\begin{document}

\newtheorem{example}{Example}

\newcommand{\nb}[2]{
    \fbox{\bfseries\sffamily\scriptsize#1}
    {\sf\small\textcolor{red}{\textit{#2}}}
}
\newcommand\ag[1]{\nb{AG}{#1}}
\newcommand\tp[1]{\nb{AA}{#1}}
\newtheorem{puzzle}{Puzzle}

\title{MCP-Orchestrated Multi-Agent System for Automated Disinformation Detection}




 \author{\IEEEauthorblockN{Alexandru-Andrei Avram, Adrian Groza and Alexandru Lecu}
 \IEEEauthorblockA{\textit{Computer Science Department of the Faculty of Automation and Computer Science} \\
 \textit{Technical University of Cluj-Napoca},
 Cluj-Napoca, Romania \\
 European University of Technology (EUt+), European Union\\
 \tt Alexandru.Avram@student.utcluj.ro, Adrian.Groza@campus.utcluj.ro}
}

\maketitle

\begin{abstract}
The large spread of disinformation across digital platforms creates significant challenges to information integrity. 
This paper presents a multi-agent system that uses relation extraction to detect disinformation in news articles, focusing on titles and short text snippets. 
The proposed Agentic AI system combines four agents: (i) a machine learning agent (logistic regression), (ii) a Wikipedia knowledge check agent (which relies on named entity recognition), (iii) a coherence detection agent (using LLM prompt engineering), and (iv) a web-scraped data analyzer that extracts relational triplets for fact checking. 
The system is orchestrated via the Model Context Protocol (MCP), offering shared context and live learning across components. 
Results demonstrate that the multi-agent ensemble achieves 95.3\% accuracy with an F1 score of 0.964, significantly outperforming individual agents and traditional approaches. 
The weighted aggregation method, mathematically derived from individual agent misclassification rates, proves superior to algorithmic threshold optimization. The modular architecture makes the system easily scalable, while also maintaining details of the decision processes.
\end{abstract}

\begin{IEEEkeywords}
Disinformation, Misinformation, Agentic AI, Model Context Protocol (MCP), Relation Extraction (RE), LLM
\end{IEEEkeywords}



\section{Motivation}

    The manipulation of information has always been a powerful tool for influencing opinions, controlling narratives, and securing power. A key point in defining and finding fake news is understanding the reasons, history or context behind this issue.
    Throughout time, altered information has been used by rulers to gain power, damaged reputations, been used for satire, or made people (and media) money.
    False and especially flashy information has always spread quickly. This truth has been observed many times, one example being author Terry Pratchett, who stated: "A lie can run 'round the world before the truth has got its boots on."
    
    In the study of information disorders, "wrong" or "false" information is generally split into three  categories, \textit{disinformation} (false information shared deliberately to mislead), \textit{misinformation} (inaccurate information shared without the intent to deceive) and \textit{propaganda} (biased or misleading information used to promote an agenda). When talking about misleading content, these three areas are treated as separate, distinct cases of information manipulation. 
    While the three categories widely differ in veracity and user intent, in the context of detecting false content, all three have to be taken into consideration, since there are chances that data in any of these classes might be skewed or fake. Moreover, from a computational perspective, they all obey the same constraints, being bound to language patterns, context, or source credibility.
    
    False information is often based on manipulating or fabricating new relationships between entities (e.g. people, organizations, events) in order to deceive audiences. 
    Relation Extraction (RE) 
    identifies and classifies semantic relationships between entities mentioned in text. 
    We aim to combat disinformation by applying RE to news articles, focusing on titles and short texts such as snippets or leading paragraphs. 
    These types of texts are 
    important because they are often the main tool for grabbing attention and swaying public opinion, since they are short, easy to understand for everybody and usually summarize the contents of articles or other types of publications.

\section{Related Work}

 

    Various technologies have been recently used to handle disinformation including Graph Neural Networks for text classification~\cite{wang2024graph},  
    LLMs and ontologies for extracting causal relationships~\cite{lecu2024using}, temporality-Aware Fake News Detection~\cite{kim2024revisiting}, information retrieval models~\cite{ebadi2021memory}, or agentic approaches for fact checking~\cite{li2024large},~\cite{horneagentic},~\cite{hou2025model}.

    Recognizing the limitations of traditional deep learning and machine learning models in capturing complex relationships in text and other contextual data, Wang et al.~\cite{wang2024graph} provide a broad overview of Graph Neural Networks (GNNs), specifically applied to text classification. They categorize GNN-based approaches into two primary types: corpus-level and document-level. Corpus-level methods build graphs for entire collections of texts to capture overall structure and relationships across documents, while document-level methods create graphs for single documents, in order to focus on how words relate to each other within that specific text. Additionally, commonly used datasets are discussed, along with evaluation metrics, while creating a comparative analysis of various GNN techniques, highlighting their strengths and limitations. 
    
    
    Lecu et al.~\cite{lecu2024using} delve into the application of Graph Neural Networks (GNNs) in text classification tasks. They introduce a unique approach that constructs graphs where nodes represent textual units (such as words or documents), and edges serve as the relationships between these units. By applying GNNs to the resulted graphs, the model can effectively learn both local and global contextual information, thus improving text classification performance. The authors demonstrate that this method outperforms traditional models, particularly in handling complex textual data with complex interdependencies.
    
    
    Various methods to detect misinformation early have been developed as a result of recognizing that timing of detection is critical to real-world mitigation. Kim et al. \cite{kim2024revisiting} present a temporality-aware evaluation framework that explicitly covers the "earliness of engagement" (how soon models can classify news after user interaction begins). They show that traditional evaluation metrics can overestimate model performance by ignoring time constraints and show that many state-of-the-art models degrade significantly in real scenarios. Their architecture aggregates what are called "earliness patterns" in order to use these results for reweighting the model. Moreover, after the reweighting takes place, a special process called "noisy edge suppression" is applied to the graph of the model, in order to identify and down-weight certain connections (edges), so as to reduce their influence in the social graph.
    
    
    Ebadi et al.~\cite{ebadi2021memory} propose a two-stage system for checking or detecting misinformation, which can be used a larger corpus of fake and conspiracy information, as opposed to most such detection systems, which can only function inside of their knowledge base, because of the high complexity of text-based machine learning. The main idea presented in the article consists of a pipeline which simulates the real-world fact-checking process, starting with a lightweight information retrieval process, used to filter unrelated articles, after which, a more complex end-to-end memory network (MemN2N) is used to return a verdict. The model is evaluated on the Fake News Challenge (FNC-1) dataset and later tested on real-world data from Snopes.com. Their system achieves performance comparable to modern models, while maintaining much lower training and detection costs, thanks to the IR component, which significantly reduces the computational load.
    This article has significantly influenced the direction of the proposed solution, moving the research towards the detection of disinformation on outside knowledge, which is not contained directly in any of the training datasets.
    
    
    Identification of fake news has generally been split into two categories as content-based and evidence-based approaches. Content-based approaches analyze text patterns, writing style, or tone in news articles, employing models such as Convolutional Neural Networks (CNNs), and transformer models such as BERT. Such models are effective in detecting linguistic cues towards misinformation. By comparison, evidence-based methods confirm claims by checking against external information, such as web pages or knowledge graphs.
    
    Li et al.~\cite{li2024large} have considered applying large language models (LLMs) for detecting fake news, particularly using few-shot and zero-shot learning. Prompt techniques, like chain-of-thought (CoT) reasoning, are also mentioned as considered effective at augmenting LLM-based classification. 
    Usually, LLMs are considered rigid, prompt-driven machines that are typically dependent on supervised data. On the other hand, agentic LLM systems have demonstrated huge potential by mimicking human expert-level thinking. These systems break down tasks into subtasks accomplished by each LLM agent. FactAgent, the solution proposed in the paper, enforces a hierarchical process where the LLM leverages internal and external knowledge in an orderly manner.
    Unlike dynamic-planning autonomous agents, FactAgent leverages a domain-optimized process without needing fine-tuning or labeled data, whilst being very adaptable to changes or new data.
    
    
    Horne et al.~\cite{horneagentic} present architectures and design patterns suitable for use in multi-agent applications. 
    The paper presents both conceptual models regarding how classic architectures (such as the layered architecture) could be integrated with agents and specific design patterns: Single Agent, Supervisor Agent, Hierarchical Agents and Agentic Network.
    
    
    AI agent orchestration, despite parallelization and task breakdown, can get computationally heavy very quickly. Coupling this issue with the number and size of the agents and these systems rapidly become large and hard to mentain.
    Hou et al.~\cite{hou2025model} explore the way in which Model Context Protocol (MCP) improves the way agents are orchestrated and used, thanks to a unified way of calling APIs and interacting with external agents and file systems. This type of implementation provides higher scalability, due to decoupling the agents from the system, while also moving the agent configuration work away from the developer, since all entities interact through the same data structures.
    Moreover, MCP offers the option to have a human-in-the-loop or supervised system, allowing power users to directly feed updates to parts of the pipeline. Also, MCP streamlines the process of agent selection. In the Model Context Protocol, each request first receives a response listing the available tools which can be used to solve the problem, out of which the orchestrator can pick which to use.

\section{Used Datasets}
    
    %
    
    The two main used datasets are "Fake News Detection" dataset from Kaggle~\cite{kaggle-dataset}, and "Snopes" fact checking dataset~\cite{vo2020facts}. 
    These datasets complement one another and enhance the data quality of the system. 
    
    The Kaggle "Fake News Detection" dataset offers a collection of labeled news articles spanning various topics and sources, providing a broad range of different misinformation types.  However, a lot of the news in this collection are specific to the USA, being less helpful for international texts.
    
    The Snopes fact-checking dataset brings in other qualities, incorporating real-world fact-checking scenarios with detailed class and domain separation of the information. This dataset is particularly valuable because it captures the reasoning processes of human fact-checkers, not just patterns of news which have been proven as fake over time, and splitting them in true statements, context-dependent claims, and distinguishing between misleading and false information.
    
    
    The web article recommendation system relies on a comprehensive domain-level evaluation system built upon the Iffy News project dataset~\cite{iffy-news}, stored as \textit{iffy-news.csv}. This curated dataset provides a great resource for filtering online information sources by reliability, by sorting domains based on their documented history of publishing misinformation, disinformation, or unreliable content.

    The training data is a mostly balanced dataset (approximately 48\% real news and 52\% fake news) composed of pre-filtered articles, which has been processed and labeled appropriately. It represents roughly 10\% of the initial data, separated before running any training.






    \section{Agentic AI fake news detection module}

\begin{figure}
        \centering
        \includegraphics[width=0.49\textwidth]{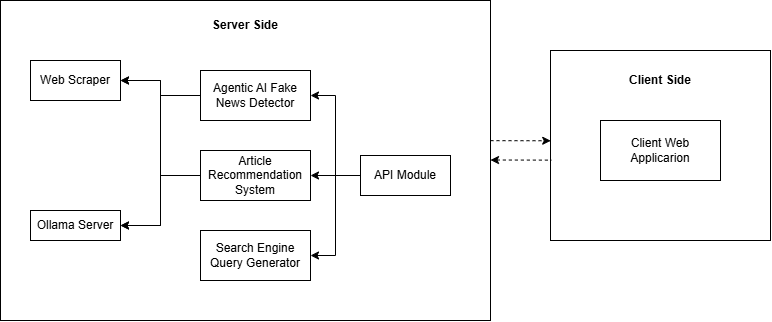}
        \caption{Main Module Diagram}
        \label{fig:main_diag}
    \end{figure}

 The solution has three parts: 
    (1) the fake news detection module based on Agentic AI; 
    (2) an article recommendation system, based on custom intelligent ranking systems (i.e. vector cosine similarity and k-nearest-neighbours); 
    (3) search engine query generator. 
    The first module is autonomous, while the second and the third modules allow human-in-the-loop. 
    These modules rely on a web scraper and on the \textit{Ollama server} \label{def:ollama-server}, which is a service used to locally host the Llama3 large-language-model. (see Figure~\ref{fig:main_diag}). 
    
\begin{figure}
            \centering
            \includegraphics[width=0.49\textwidth]{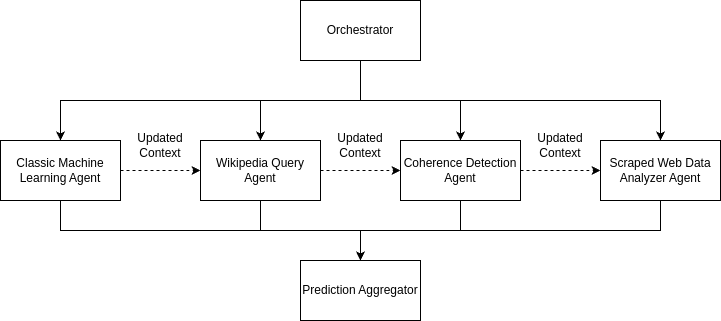}
            \caption{Agentic AI Fake News Detector Module Diagram}
            \label{fig:agentic_diag}
        \end{figure}

    The modules consists of four agents,
     accompanied by two auxiliary modules, the \textit{orchestrator}, which helps in defining the pipeline, and the \textit{aggregator}, which cumulates the agents results into a single output (Figure~\ref{fig:agentic_diag}). 
Technically, each agent can also function as a standalone service, or can be integrated by other modules.     
        
All agents communicate with the Ollama Server, when semantically complex tasks are managed, and with the Web Scraper, in order to create a unique pool of information as a starting common truth source. As the agents run, each agent has its input to the existing context, updating it and passing it further to the next agents in the pipeline.


    \subsection{Classic Machine Learning Agent}
       The aim is to provide fast prediction on input, with the ability to adapt its parameters, as new information becomes available.
        The agent is based on a combination of a \textit{Hashing Vectorizer} and a \textit{Stochastic Gradient Descent (SGD) classifier}. 

        
        \textit{Stochastic Gradient Descent} 
        minimizes loss by updating its parameters incrementally. 
        Here, the logistic regression is used, which outputs probabilities for its result. These probabilities are used in the given task to classify the user input as factual or false and to calculate the confidence score of the agent.  
        Since SGD is fit for high-dimensional datasets such as entity vectorized text, it is adequate for fake news where there is no specific predefined structure, and inputs can widely differ. To prevent overfitting, L2 is used as regularisation method. 
        
        
        \textit{Hashing Vectorizer (HV)} 
        is both memory-efficient and scalable. 
        Unlike other vectorizers (such as Count or TF-IDF), the Hashing Vectorizer does not require a set of words and does not store feature names, which drastically reduces memory usage. This is highly relevant in the high RAM usage context it is being run in. 
        From Roshan et al.~\cite{roshan2023comparative}, we learn that generally, TF-IDF Vectorizer performs better than a Hashing Vectorizer for fake news detection. 
        However, from their findings, HV specifically has similar if not better performance for the methods which were tackled in the testing phase (e.g. logistic regression, neural networks training, support vector machines or gradient boosting) and our desired metric (i.e. accuracy).
        Additionally, HV allows for streaming data, enabling a continuous learning cycle. 
        Upon finding new information, the model can be updated on the go in its current running session. 
        This agent uses the \textit{context} of the Web Scraper Module in order to obtain new data based on user input, which it then uses to update its own weights before making a prediction (i.e. online learning), thus leveraging the power of the internet as well.

\begin{example}
    The model is trained on a medical dataset, but need to be used on political texts. The system can fetch new information from the internet and use it to update its weights to better classify the novel information.
\end{example}
        
        
        
        
        Technically, there are two components: (i) the trainer - responsible for the initial model creation, training, evaluation, and (ii) the agent runner - runtime behavior of the agent, including inference, online updates, and integration with scraped or context data



           \begin{table}
            \caption{Agent Performance Metrics}
            \label{tab:agent_metrics}
            \centering
            \begin{tabular}{lrrrr}
                Agent & Precision & Recall & Accuracy & F1 \\
                \hline
                Classic ML & 1.00 & 0.22 & 0.61 & 0.36 \\
                Wikipedia Query & 0.68 & 0.64 & 0.67 & 0.66 \\
                Coherence Detector & 0.79 & 0.38 & 0.64 & 0.51 \\
                Scraped Data Analyzer & 0.97 & 0.78 & 0.88 & 0.86 \\
                Weight Aggregate (best run) & 1.00 & 0.90 & 0.95 & 0.947 \\
            \end{tabular}
        \end{table}
    
        This agent is efficient when testing data is in the same domain or knowledge area as the training set. 
        As~\autoref{tab:all_matrices} shows, the Classic Machine Learning model achieved up to 92\% accuracy (42\% True positives + 50\% true negatives) 
        on known or partially known information (i.e. the model has all or most of the necessary information to infer an accurate output from its training data). However, the model quickly falls short when applied to out-of-domain data or scenarios which have not been encountered during training. ~\autoref{tab:all_matrices} also shows that the same model's performance degrades dramatically when met with unfamiliar data distributions, achieving only 61\% accuracy, on the same test data, with 39\% false negative rate, which is a specifically bad area of the confusion matrix.
        
        \begin{table}
            \caption{Known Data Confusion Matrix}
            \label{tab:all_matrices}
            \centering
            \begin{tabular}{lrrrr}
                
               Agent & True + & True - & False +  & False - \\ \hline
               Classical ML (Known Data) & 42 & 50 & 8 & 0 \\ 
               Classical ML (Unknown Data) & 11 & 50 & 39 & 0 \\ 
               Wikipedia Data & 32 & 35 & 18 & 15 \\ 
               Coherence Detection & 19 & 45 & 31 & 5 \\ 
               Scraped Data Analysis & 39 & 49 & 11 & 1 \\ 
               Weight Aggregate (best run) & 45 & 50 & 5 & 0 \\

            \end{tabular}
        \end{table}
        
    
    \subsection{Wikipedia Knowledge Agent}
            The main idea behind this agent is that a veracity benchmark can be created by checking how well ideas and terms match up with already known, widely accepted knowledge and data. 
            The agent relies on Named Entity Recognition (NER) and Relation Extraction (RE), and it runs as a combination between a fact-checker and a semantic comparator. 
            
            Agent confidence can be calculated as keyword overlap between the input text and retrieved summaries, assuming that real news align more closely with encyclopedia knowledge than false claims do. This overlap is computed using word sets which are processed through lemmatization and selected as the most relevant parts of speech in most sentences, capturing the main topics of the text rather than broader information.
        \begin{example}[Raw data] Let the user input: \textit{Marie Curie discovered radium and won two Nobel Prizes}, for which Wikipedia-fetched information $W$ is 
                  (1) \textit{Marie Curie was a physicist and chemist who conducted pioneering research on radioactivity. She was the first woman to win a Nobel Prize[...]}, 
                  (2) \textit{Radium is a radioactive element discovered by Marie and Pierre Curie in 1898. It was formerly used [...]}, 
                  (3) \textit{The Nobel Prizes are international awards given annually in several categories including physics and chemistry. Marie Curie received the Nobel Prize in Physics in 1903 and in Chemistry in 1911.[...]}.
Extracted search terms, normalized:
\begin{tabular}{lp{7.cm}}
        $In$: & ["Marie Curie", "win", "discover", "Radium", "Nobel Prize"]\\
$W$: & ["Marie Curie", "physicist", "chemist", "conduct", "research", "radioactivity", "win", "Nobel Prize", "discover", "Radium"]\\
$In \cap W$: & ["Marie Curie", "win", "discover", "Radium", "Nobel Prize"]
     \end{tabular}       
Since $In \cap W$ contains the same terms as user input, there is a full overlap. Hence, the verdict is real information.

        \end{example}
Formally, the algorithm starts by extracting the part of speech
$POSs = \{\text{NOUN, PROPN, VERB}\}$.
Then, the keywords from the input text $t_1$ are collected             
in $K_1 \;=\; \bigl\{\ell(w)\mid w\in\text{t}_1,\;\mathrm{POS}(w)\in POSs\bigr\}
$, while the keywords from the wiki in $K_2 \;=\; \bigl\{\ell(w)\mid w\in\text{t}_2,\;\mathrm{POS}(w)\in POSs\bigr\}$. The overlap score $O$ is given by 

   $$ O =\frac{\bigl|\,K_1 \cap K_2\,\bigr|}{\max\bigl(|K_1|,1\bigr)}, $$

\noindent
where \(w\) is a word token, \(\ell(w)\) its lemma, and \(\mathrm{POS}(w)\in\mathrm{POSs}\) restricts the set to the selected parts of speech.

            
            The extracted terms are used to gather short Wikipedia summaries, which serve as a reality check against general knowledge, using the following workflow:
        First, NER extracts entities like \textit{person, organization, geopolitical entity, event or work of art}, as well as nouns and proper nouns. 
            Second, these entities are queried to Wikipedia using a specialized function. 
             Third, the answers are used to compute the keyword overlap.
           
            This simple workflow aligns with key goals in disinformation detection: being transparent, using trusted sources, tracking search terms and keyword matches, and remaining up to date as knowledge sources evolve. However, it has certain downsides: it relies on surface-level text comparison and article summaries, which might miss details or very new information. Still, it offers a practical starting point for fact-checking applications, especially when coupled with other types of AI agents in a larger system.

        \begin{figure}
                \centering
                \includegraphics[width=0.5\textwidth]{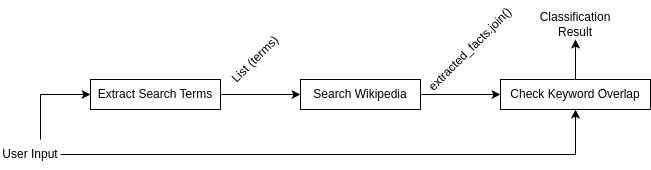}
                \caption{Flow of the Wiki Agent algorithm}
            \end{figure}
            The Wikipedia data agent is efficient at retrieving factual information on subjects. 
            It is limited to the articles available on the website, or it can misfire in certain cases when sarcasm is involved or when sketchier text types are analyzed such as propaganda or conspiracies. Line {two} in~\autoref{tab:all_matrices} indicates a moderate performance with an overall accuracy of 67\% (32\% + 35\%). 
            Unlike the classic ML agent, this agent shows a more balanced error distribution, with both false positives and false negatives, thus showing a behavior closer to naturally occurring error patterns.
            
    
    \subsection{Coherence Detection Agent}
            
            After evaluating two datasets: ISOT, a large scale political news, and HWB, on health and well-being dataset, Singh et al.~\cite{singh2020coherence} observed that "fake news articles exhibit lower textual coherence, as compared to legitimate news". 
            Coherence is a useful test of how intelligible a piece of text is overall. When a news article jumps all over the place from one unrelated thing to another, lacks a certain direction, or feels disjointed, it may not just be bad writing, but it may even suggests intentional lying, manipulation, or bad intent in reporting, all of which point to a case of fake news. 
           For coherence detection, we employ here a prompt engineering approach. 
        
            The Coherence Detection Agent is a 
            LLM-based module designed to grade text coherence. 
            That is whether a paragraph or sentence is logically and chronologically organized, semantically correct, and syntactically sound and structured. 
            The evaluation task is delegated to the LLM running in the Ollama Server (~\autoref{def:ollama-server}), rather than processing text using NLP. As Liu et al. show~\cite{liu2024unlockingstructuremeasuringintroducing}, LLMs might be the better choice when compared to traditional NLP methods for coherence detection because they understand language in different ways.
            
            The "Act As"~\cite{act-as-pattern} pattern has been used in order to make the model act as an assistant helping with coherence detection.
            The aim of this pattern is to improve the language model's predictions by narrowing their scope through the queried prompt. 
            Here, the prompt has been reformulated in order to set a context before putting the task forward.
                   
        
            The Coherence Detection agent displays mixed performance, \autoref{tab:agent_metrics} showing an overall accuracy of 64\%. In this case, it is normal to have rather low accuracy in classifying texts as factual or not, since coherence is only a smaller part of disinformation detection. Pieces of news can be factual but poorly written, while fake news can also be coherent, despite not containing reliable information.
            
    
    \subsection{Web Scraped Data Analyzer Agent}
            We used the Llama3 8 billion version, 
            which is a  decoder-only transformer architecture, built out of one embedding layer, 32 transformer layers and a dense layer~\cite{meta-llama}~\cite{medium-llama}.
            Effective detection models rely heavily on large volumes of data to learn patterns and accurately identify misinformation~\cite{khder2021web}.
            
            The web scraping pipeline is based on a triplet extraction system, which is split into logical stages: data fetching, data preprocessing, data comparison: 
First, data Fetching querying the web for relevant content.
            Second, SpaCy's English core transformer relation extraction model is used. 
        Third, the two sets of triplets are compared.  A structured prompt is used, containing the initial user claim, triplets extracted from the claim, and triplets extracted from the web data, along with details about how the LLM should structure the output.
         Finally, the Scraped Web Data Analyzer Agent returns a structured output containing a label (real/fake), a confidence level, and an explanation of its reasoning.
            
          \begin{example}[Generated explanations - Coherence Agent]
            "The paragraph contains a statement that appears to be a factual assertion, which is coherent and makes sense."
          \end{example}

            The Web Scraped Data Analyzer agent has the best single-agent performance out of the four agents (\autoref{tab:agent_metrics}). The 88\% accuracy will play a big role in deciding the model's response weight in the final result, and is a direct result of the agent's real-time access to the internet and to the Llama3-8B model. 
            

    \section{Agent orchestration and Human-in-the-loop}
    \subsection{Agent Orchestration and Aggregation}
        Since the system has a multi-source, multi-agent architecture, the agents and data need to be \textit{orchestrated} 
        so that: (i)  no context is lost, and (ii) each agent has an extra input, making the new information available help the other agents. 
        
        The orchestrator is the first sub-module accessed in the Agentic AI Fake News Detector (Figure~\ref{fig:agentic_diag}), and calls for the execution and order of execution of the agents.
        A configuration which uses the LangChain framework is defined, structured as a sequence of data analyzers. The orchestration is based on the Model Context Protocol (MCP), and coordinates multiple submodules that share a common context, updating it as each of them runs. The pipeline is built using \textit{RunnableLambda} operators, enabling modular and asynchronous component chaining, very important for system expansion possibilities.
        
        The \textit{Aggregator} represents the final layer in the Agentic AI module orchestration. It consolidates the results of all prior sub-modules into a unified output. 
        This component implements heuristics and confidence weighting to infer a final decision or summary.
        It functions by calculating a weighted average of the results of each agent, while also taking into consideration the confidence of each agent's answer.
        The weights considers each agent's misclassification rate. 
        Initially, weights were obtained by exploring different value combinations, via . 
        After updating the weights, each agent has the following relevance in the calculation of the final result: 
        Classic Machine Learning Agent - 19\%,
            Wikipedia Query Agent - 23\%,
            Coherence Detection Agent - 17\%,
            Scraped Web Data Analyzer Agent - 41\%.

\begin{figure}
            \centering
            \includegraphics[width=0.5\textwidth]{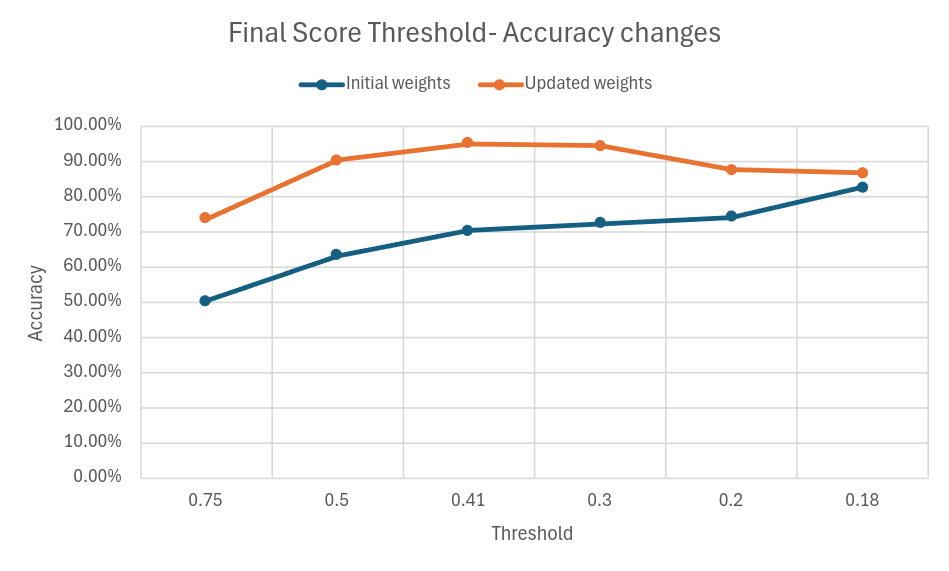}
            \caption{Accuracy Fluctuation by Threshold}
            \label{fig:threshold_acc}
        \end{figure}

        Calculating the threshold and weights significantly improved the results when compared to the algorithmically obtained weights - by looping through different values for each threshold. 
        After misclassification was calculated for each agent it is subtracted from 1 to yield the percentage of correctly classified results. These percentages were then used to determine the weights. Specifically, the weight \( w_i \) for each agent \( i \) was computed as:
        
        \begin{equation}
            w_i = \frac{c_i}{\sum_{j=1}^{n} c_j}
\end{equation} 
where \( c_i = 1 - m_i \), and \( m_i \) represents the misclassification rate of agent \( i \).
         As Figure~\ref{fig:threshold_acc} demonstrates, the weights found through the mathematical method perform significantly better, reaching a peak of 95.20\% accuracy for threshold 0.41.

        There are also \textit{logical explanations} for why these percentages are low/high depending on agent. 
        The coherence detection agent has the lowest say in the final result, since the text is preprocessed, and slightly reformulated as well before the agents even kick in to do the classifications. The classic machine learning agent has the lowest percentage out of the fact checking agents, because it has access to the least amount of updated data, despite updating its network on web scraped data as well. Old pre-trained data might also interfere with the node updates sometimes, leading to skewed results. The Wikipedia query agent and scraped data analyzer have the highest weights because they have access to the biggest amount of data, which is also up-to-date or almost up-to-date at the moment of request.
    
    \subsection{Human-In-The-Loop sub-system}
       \begin{figure*}
            \centering
            \includegraphics[width=0.85\textwidth]{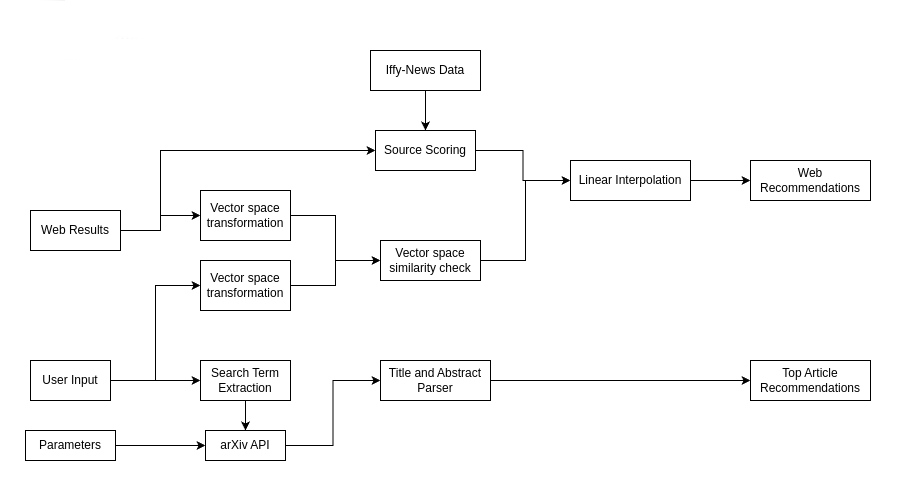}
            \caption{Article Recommendation System}
            \label{fig:recomm_diagram}
        \end{figure*}
        
        The Human-in-The-Loop sub-system 
        is a two-track article recommender: (1) one side tailored for selecting relevant web-based sources, and (2) the other for retrieving scientific literature, specifically from the arXiv database (\autoref{fig:recomm_diagram}).

        
        First, the \textit{Web Recommendations} module ranks relevant and trustworthy web sources in relation to a user’s input. 
        This part of the system operates on top of external search engine ranking. 
        It applies its own scoring by integrating semantic similarity and source trustworthiness checking, as follows:
        
            \begin{enumerate}
                \item \textit{Scoring Source Reputation}. To address the issue of information integrity, even before checking how relevant articles are to the input text, each piece of information is also assigned a source credibility score, computed using a domain-level heuristic based on the Iffy News dataset.
                \item \textit{Extracting Sentence Embeddings}. The main actor in the extraction process is a pretrained sentence embedding model. This model is optimized for efficient encoding of sentences and short documents into dense vectors.
                \item \textit{Calculating Semantic Similarity}. Once sentence embeddings are obtained for both the user input and each article, the system calculates pairwise semantic similarity using cosine similarity. 
                \item \textit{Calculating Combined Scoring}. The final score for each article $k$ is computed using a linear interpolation of the semantic similarity $sim_k$ and the source credibility:
                $score_k = \alpha \cdot sim_{k} + (1 - \alpha) \cdot source_k$, 
                                where $\alpha \in [0,1]$ 
                                allows the system to prioritize semantic relevance while still using the source reliability score as well.
            \end{enumerate}
        
        
            Second, the \textit{Scientific Article Recommendations} module 
            focuses on academic sources. 
            While the web offers quick and varied perspectives, scientific articles offer a different approach to information: they bring in rigor, structure, and an evidence-based view of the world. 
            At the core of the system lives the \textit{arXiv API}, which provides access to a massive collection of scientific papers in a wide range of domains, such as physics, computer science, mathematics etc. Rather than solely relying on keyword searching, the module formulates a special query, narrowing its word matching system to titles and abstracts, which tend to encapsulate the main idea of a paper. The results returned by the arXiv API are delivered in Atom XML format, a structured but verbose structure, from which the system extracts only the relevant information: the paper’s title, abstract, authors, and a stable link to the arXiv page of the article (which is leveraged to link to the direct article).

\section{Running Experiments}
            In the context of fake news detection, TF-IDF often struggles to make reliable predictions when dealing with content outside its training data. 
            The model places heavy emphasis on individual words, without considering the broader context in which they appear. 
            This means that even a single word, if strongly associated with false information during training can skew the classification of an entirely new and unrelated piece of content. As a result, the presence of a known "suspicious" word in an unfamiliar context can lead the model to incorrectly flag accurate information as fake, simply because it lacks the ability to understand nuance or semantics.

    \subsection{LLM Fine-Tuning Approach}
            The HuggingFace Transformers library has been used along with PyTorch, in order to fine-tune a large language model on the dataset. 
            Three different fine-tuning methods have been used: Low-Rank Adaptation (LoRA), Bias Fine-Tuning (BitFit) and Knowledge Distillation (KD).
        
            The model used for the following experiments is called "DistilBERT base model (uncased)" and has been developed by HuggingFace as a distilled version of google-bert/bert-base-uncased. It is a light model, which removed 40\% of its predecessor's parameters, while retaining 97\% of its performance and being significantly faster~\cite{sanh2019distilbert}. 
            This model has been chosen, since its primarily aimed at being fine-tuned to make decisions (classification problems).
        
                Low-Rank Adaptation (LoRA) fine-tunes a LLMs by minimizing the number of modified parameters, to improve efficiency. As model size grows, it becomes less feasible to change the entire weight matrix of the model when fine-tuning~\cite{hu2021lora}.
                Low-Rank Adaptations functions by freezing the original weight matrix in its entirety (making it unmodifiable) and updating 2 smaller matrices which represent the decomposition of the original weight matrix. Given the size of the weight matrix as being \(d \times d\), this update can be formulated as $\Delta{W} = A B$,
              where matrix \textit{A} contains the linearly independent "bias" vectors (\( A \in \mathbb{R}^{d \times r} \)), and matrix \textit{B} contains the information necessary to rebuild the complete and updated weight matrix (\( B \in \mathbb{R}^{r \times d} \)).
                Hence, the updated weight matrix has the structure:
                                            $W' = W + \Delta{W} = W + A B$, 
                where \(\Delta{W} \in \mathbb{R}^{d \times d}\).
                LORA is efficient when fine-tuning, while also keeping learning times relatively low and not needing additional resources (i.e. other models already running). 
                

                \begin{table}
                    \caption{Fine Tuning Methods Confusion Matrices}
                    \label{tab:all_matrices_FT}
                    \centering
                    \begin{tabular}{lrrrr}
                        
                       Agent & True + & True - & False +  & False - \\ \hline
                       Low-Rank Adaptation & 47.68 & 52.28 & .02 & .01 \\ 
                       Bias Fine-Tuning & 47.61 & 51.45 & .10 & .85 \\ 
                       Knowledge Distillation & 47.69 & 52.27 & .01 & .02 \\

                    \end{tabular}
                \end{table}
            
                The BitFit LLM fine-tuning method is a simple and efficient way of specializing transformer-based language models. In a very similar fashion to Low-Rank Adaptation, it aims to modify only a small part of the weight matrix in order to obtain the desired results. As opposed to LoRA, the changes happen directly in the original weight matrix, as opposed to a separate one which contains only the modified weights. BitFit focuses on updating \textit{bias terms} (i.e. terms which will make the network favor certain answers) in specialized areas~\cite{zaken2021bitfit}
                Bias Fine-Tuning operates by freezing all of the model's parameters, except for the bias terms. This is an efficient way of training since, in most models, the majority of the parameters are located in the weight matrix, \textit{W}, while the bias vector \textit{b} is very small in comparison. Therefore, given the equation
            $h=Wx+b$, $W$ is fixed (frozen) and $b$ is updated using: 
            
                \begin{equation}
                   h=Wx+(b+\Delta{b}); \Delta{b} \in \mathbb{R}^{d \times d}
                \end{equation}
                where \(\Delta{b}\) is a small, learnable adjustment, which captures the specific details of the fine-tuning.
                
                While similar in training time and results to LoRA, Bias Fine-Tuning yields slightly worse results, both as false positives and especially false negatives (lines 1 and 2 in~\autoref{tab:all_matrices_FT}).
            
                
                Knowledge Distillation works by transferring soft targets (i.e. the probability distribution over the output classes) from the teacher model to the student model, instead of directly training the student model on labeled data.
            Knowledge Distillation has had the best performance over multiple test runs (~\autoref{tab:all_matrices_FT}, third row), however it is way more resource-heavy, both as storage and computing power.
    
    \subsection{Agentic Approach}
        
            For the bare model experiments, \textit{distilbert-base-uncased-finetuned-sst-2-english} 
            (already fine-tuned model for the english language) has been used in single mode, while in ensemble mode, it has been coupled with \textit{roberta-base}, to improve the predictions.
             The 2-agent with finetuned model version uses a version of \textit{roberta-base}, specifically fine-tuned on a knowledge graph derived from a training dataset.
        
                Initially, the accuracy of a single LLM was evaluated to establish a baseline for further improvement. The results from this single-agent setup were notably poor, with the model displaying almost random behavior and a high rate of incorrect classifications. The most critical area is that of false negatives, where fake content was incorrectly identified as genuine.

                \begin{table}
                    \caption{Agent Ensembles Confusion Matrices}
                    \label{tab:all_matrices_FT_ensemble}
                    \centering
                    \begin{tabular}{lrrrr}
                        
                       Ensemble Type & True + & True - & False +  & False - \\ \hline
                       Single Model & 7 & 43.5 & 34 & 15.5 \\ 
                       2-agent Ensemble & 0 & 59 & 41 & 0 \\ 
                       Fine-Tuned Agent Ensemble & 46.5 & 53.25 & 0 & 0.25 \\

                    \end{tabular}
                \end{table}
                
                
                To address these shortcomings, a second model was added, forming a two-agent ensemble. Each agent independently evaluates the input and produces a result. The final classification is derived by averaging the outputs of the two models. This approach shows a clear improvement over the single-agent baseline, most notably, it eliminating false positives (real content is rarely misclassified as fake).
                
                However, this solution introduces a new problem. The ensemble tends to overcorrect by classifying all inputs as fake. This aggressive filtering successfully addresses false positives, but at the cost of increasing false negatives — an already existing problem of the single-agent model. This indicates that while the ensemble strategy for untrained models improves certain aspects of the system, it does not offer an optimal solution, leaning toward excessive skepticism in classification.
                
                
                To improve classification performance, a fine-tuned model has been introduced into the agent ensemble. 
                This model was trained on a curated dataset, which had been filtered and aggregated beforehand, in order to contain multiple sources of data. 
                The results of this change are very noticeable.
                
                
                Compared to the previous configurations, this version of the system has greately increased performance. Misclassifications are almost entirely eliminated, with both false positives and false negatives reduced to a minimum. The model now indicates a far more accurate and consistent classification of the input data. While earlier versions tended to either behave randomly (in the single-agent case) or highly overcorrect by labeling everything as fake (in the simple, bare-model ensemble), the inclusion of the fine-tuned model brings an improved decision-making capability.
                
                The results suggest that fine-tuning with domain-specific data plays a critical role in enhancing model reliability. It helps the system understand the small details that general models miss. Instead of only relying on averaging the outputs of two agents which led to unbalanced and inaccurate results, adding a trained, task-specific model makes the system more reliable.
    
\section{Discussion}



The proposed Agentic AI system achieves 95.3\% accuracy and an F1 score of 0.964. However, several practical and theoretical challenges deserve attention.

The system's dependence on web-sourced data creates both opportunities and risks. Although real-time access to information through the DuckDuckGo API helps to detect new disinformation patterns, if the API fails or search results are manipulated, the system's effectiveness drops. 
More concerning is the recursive problem where disinformation might influence the search results used to detect it. This suggests the need for multiple fallback data sources and more robust verification methods that do not depend solely on web searches.

Performance in different knowledge domains remains an important area for investigation. While our evaluation used mixed datasets covering various topics, we did not systematically analyze performance variations by domain. Different types of content—political news, scientific articles, health information, or entertainment news—likely present unique challenges. Political content may contain more opinion-based statements, while scientific articles rely heavily on technical terminology and citations. The Wikipedia agent, for instance, would presumably perform better for topics with comprehensive encyclopedia coverage but may struggle with emerging events or highly specialized technical fields. 

The bias in training data and reference sources poses a subtle challenge. 
The Wikipedia and Web search results reflect existing societal biases that vary by topic and region. 
This might be problematic in case of global disinformation campaigns and the need for culturally sensitive detection methods.

The current architecture has not been extensively tested against adversarial attacks. 
Modern disinformation campaigns use techniques that include mixing facts with misleading context and coordinating across platforms. Static weight aggregation, despite mathematical optimization, may prove vulnerable to attackers who study and exploit the system's patterns. Additionally, the temporal nature of truth presents challenges not fully addressed in our implementation. Information valid at one time may later prove false, which requires systems that can track and update assessments over time.

Despite these limitations, the multi-agent approach offers clear advantages. Modular design allows improvements to individual components without complete system redesign. Using diverse methods, from machine learning to knowledge verification, provides cross-validation opportunities~\cite{lupu2023cross} against various types of disinformation. 
In line with~\cite{groza2022fact}, having multiple agents explain their reasoning improves transparency and helps human operators make informed decisions.


\textbf{Acknowledgment}. A. Groza is supported by a grant of the Ministry of Research, Innovation and Digitization, CCCDI-UEFISCDI, project number
PN-IV-P6-6.3-SOL-2024-2-0312 within PNCDI IV.

\section{Conclusion}
    Detecting misinformation through relation extraction based on ML and LLMS proves to be a highly complex one, which poses not only technical but also language and context problems.
    Agentic methods have proven to perform better in this task, since the multiple sources of different knowledge work together similar to a team of professionals with expertise in different fields, and are able to narrow down on the finer details of each article or case presented.
    
    Methods such as ML and model training 
    quickly fall short when it comes to new information or out-of-scope data.
    One takeaway from our Agentic AI solution is that complementarity is a valuable quality in such complex tasks. 
    No single technique, whether a light TF-IDF classifier or the 8-billion-parameter LLM, can carry the task of fact-checking alone.
    
    In line with the MCP, we learn that context is a "first-class citizen". Incremental learning, shared memory, and context passing allowed each agent to understand the updates introduced by its predecessors. This meant that even components like the ML model could update its weights with new relations that may have even contradicted its initial prediction.

    
    Future development may focus on several areas. 
    First, reducing API dependencies through alternative data sources would improve reliability. Second, incorporating temporal analysis would help track how information evolves. Third, expanding language support would enable detection of global disinformation campaigns. 

    Our work demonstrates that combining diverse AI approaches through MCP orchestration can achieve significant improvements in detection accuracy. As disinformation tactics grow more sophisticated, modularity and the human-in-the-loop feature and of the proposed solution positions it well for future enhancements and real-world deployment.



\bibliographystyle{IEEEtran}

\nocite{*}

\bibliography{bib}

\begin{thebibliography}{10}
\providecommand{\url}[1]{#1}
\csname url@samestyle\endcsname
\providecommand{\newblock}{\relax}
\providecommand{\bibinfo}[2]{#2}
\providecommand{\BIBentrySTDinterwordspacing}{\spaceskip=0pt\relax}
\providecommand{\BIBentryALTinterwordstretchfactor}{4}
\providecommand{\BIBentryALTinterwordspacing}{\spaceskip=\fontdimen2\font plus
\BIBentryALTinterwordstretchfactor\fontdimen3\font minus \fontdimen4\font\relax}
\providecommand{\BIBforeignlanguage}[2]{{%
\expandafter\ifx\csname l@#1\endcsname\relax
\typeout{** WARNING: IEEEtran.bst: No hyphenation pattern has been}%
\typeout{** loaded for the language `#1'. Using the pattern for}%
\typeout{** the default language instead.}%
\else
\language=\csname l@#1\endcsname
\fi
#2}}
\providecommand{\BIBdecl}{\relax}
\BIBdecl

\bibitem{wang2024graph}
K.~Wang, Y.~Ding, and S.~C. Han, ``Graph neural networks for text classification: A survey,'' \emph{Artificial Intelligence Review}, vol.~57, no.~8, p. 190, 2024.

\bibitem{lecu2024using}
A.~Lecu, A.~Groza, and L.~Hawizy, ``Using llms and ontologies to extract causal relationships from medical abstracts,'' \emph{Procedia Computer Science}, vol. 244, pp. 443--452, 2024.

\bibitem{kim2024revisiting}
J.~Kim, J.~Lee, Y.~In, K.~Yoon, and C.~Park, ``Revisiting fake news detection: Towards temporality-aware evaluation by leveraging engagement earliness,'' \emph{arXiv preprint arXiv:2411.12775}, 2024.

\bibitem{ebadi2021memory}
N.~Ebadi, M.~Jozani, K.-K.~R. Choo, and P.~Rad, ``A memory network information retrieval model for identification of news misinformation,'' \emph{IEEE Transactions on Big Data}, vol.~8, no.~5, pp. 1358--1370, 2021.

\bibitem{li2024large}
X.~Li, Y.~Zhang, and E.~C. Malthouse, ``Large language model agentic approach to fact checking and fake news detection,'' in \emph{ECAI 2024}.\hskip 1em plus 0.5em minus 0.4em\relax IOS Press, 2024, pp. 2572--2579.

\bibitem{horneagentic}
D.~Horne, ``The agentic ai mindset--a practitioner’s guide to architectures, patterns, and future directions for autonomy and automation,'' \emph{ResearchGate}, 2025.

\bibitem{hou2025model}
X.~Hou, Y.~Zhao, S.~Wang, and H.~Wang, ``Model context protocol (mcp): Landscape, security threats, and future research directions,'' \emph{arXiv preprint arXiv:2503.23278}, 2025.

\bibitem{kaggle-dataset}
\BIBentryALTinterwordspacing
{Sameer Patel}, ``Fake news detection,'' [Online; accessed 06-December-2024]. [Online]. Available: \url{https://www.kaggle.com/code/therealsampat/fake-news-detection}
\BIBentrySTDinterwordspacing

\bibitem{vo2020facts}
N.~Vo and K.~Lee, ``Where are the facts? searching for fact-checked information to alleviate the spread of fake news,'' \emph{arXiv preprint arXiv:2010.03159}, 2020.

\bibitem{iffy-news}
\BIBentryALTinterwordspacing
{The Iffy News project}, ``Reliability research,'' [Online; accessed 14-June-2025]. [Online]. Available: \url{https://iffy.news/}
\BIBentrySTDinterwordspacing

\bibitem{roshan2023comparative}
R.~Roshan, I.~A. Bhacho, and S.~Zai, ``Comparative analysis of tf--idf and hashing vectorizer for fake news detection in sindhi: A machine learning and deep learning approach,'' \emph{Engineering Proceedings}, vol.~46, no.~1, p.~5, 2023.

\bibitem{singh2020coherence}
I.~Singh, P.~Deepak, and K.~Anoop, ``On the coherence of fake news articles,'' in \emph{ECML PKDD 2020: Workshops of the European Conf. on Machine Learning and Knowledge Discovery in Databases, Ghent, Belgium, September 14--18, 2020}.\hskip 1em plus 0.5em minus 0.4em\relax Springer, 2020, pp. 591--607.

\bibitem{liu2024unlockingstructuremeasuringintroducing}
\BIBentryALTinterwordspacing
Y.~Liu, Y.~Su, E.~Shareghi, and N.~Collier, ``Unlocking structure measuring: Introducing pdd, an automatic metric for positional discourse coherence,'' 2024. [Online]. Available: \url{https://arxiv.org/abs/2402.10175}
\BIBentrySTDinterwordspacing

\bibitem{act-as-pattern}
\BIBentryALTinterwordspacing
{Stephen Redmond}, ``Llm prompt engineering patterns,'' [Online; accessed 09-June-2025]. [Online]. Available: \url{https://www.linkedin.com/pulse/llm-prompt-engineering-patterns-stephen-redmond/}
\BIBentrySTDinterwordspacing

\bibitem{meta-llama}
\BIBentryALTinterwordspacing
Meta, ``Introducing meta llama 3: The most capable openly available llm to date,'' [Online; accessed 25-April-2025]. [Online]. Available: \url{https://ai.meta.com/blog/meta-llama-3/}
\BIBentrySTDinterwordspacing

\bibitem{medium-llama}
\BIBentryALTinterwordspacing
X.~Zhao, ``Deep dive into llama 3,'' [Online; accessed 25-April-2025]. [Online]. Available: \url{https://medium.com/@zhao_xu/deep-dive-into-llama-3-351c7b4e7aa5}
\BIBentrySTDinterwordspacing

\bibitem{khder2021web}
M.~A. Khder, ``Web scraping or web crawling: State of art, techniques, approaches and application.'' \emph{International Journal of Advances in Soft Computing \& Its Applications}, vol.~13, no.~3, 2021.

\bibitem{sanh2019distilbert}
V.~Sanh, ``Distilbert, a distilled version of bert: smaller, faster, cheaper and lighter,'' \emph{arXiv preprint arXiv:1910.01108}, 2019.

\bibitem{hu2021lora}
E.~J. Hu, Y.~Shen, P.~Wallis, Z.~Allen-Zhu, Y.~Li, S.~Wang, L.~Wang, and W.~Chen, ``Lora: Low-rank adaptation of large language models,'' \emph{arXiv preprint arXiv:2106.09685}, 2021.

\bibitem{zaken2021bitfit}
E.~B. Zaken, S.~Ravfogel, and Y.~Goldberg, ``Bitfit: Simple parameter-efficient fine-tuning for transformer-based masked language-models,'' \emph{arXiv preprint arXiv:2106.10199}, 2021.

\bibitem{lupu2023cross}
D.~Lupu, A.~Groza, and A.~Pease, ``Cross-validation of answers with sumo and gpt.'' in \emph{KBC-LM/LM-KBC@ ISWC}, 2023.

\bibitem{groza2022fact}
A.~Groza and {\'A}.~Katona, ``Fact-checking with explanations,'' in \emph{2022 24th Int. Symposium on Symbolic and Numeric Algorithms for Scientific Computing (SYNASC)}.\hskip 1em plus 0.5em minus 0.4em\relax IEEE, 2022, pp. 150--157.

\bibitem{o2018cognitive}
S.~Schofield, ``Cognitive bias in clinical medicine,'' \emph{Journal of the Royal College of Physicians of Edinburgh}, vol.~48, no.~3, pp. 225--232, 2018.

\bibitem{kliegr2021review}
T.~Kliegr, {\v{S}}.~Bahn{\'\i}k, and J.~F{\"u}rnkranz, ``A review of possible effects of cognitive biases on interpretation of rule-based machine learning models,'' \emph{Artificial Intelligence}, vol. 295, p. 103458, 2021.

\bibitem{rastogi2022deciding}
C.~Rastogi, Y.~Zhang, D.~Wei, K.~R. Varshney, A.~Dhurandhar, and R.~Tomsett, ``Deciding fast and slow: The role of cognitive biases in ai-assisted decision-making,'' \emph{Proceedings of the ACM on Human-Computer Interaction}, vol.~6, no. CSCW1, pp. 1--22, 2022.

\bibitem{bracsoveanu2019semantic}
A.~M. Bra{\c{s}}oveanu and R.~Andonie, ``Semantic fake news detection: a machine learning perspective,'' in \emph{Advances in Computational Intelligence: 15th Int. Work-Conf. on Artificial Neural Networks, IWANN 2019, Gran Canaria, Spain, June 12-14, 2019, Proceedings, Part I 15}.\hskip 1em plus 0.5em minus 0.4em\relax Springer, 2019, pp. 656--667.

\bibitem{tsai2023stylometric}
C.-M. Tsai, ``Stylometric fake news detection based on natural language processing using named entity recognition: In-domain and cross-domain analysis,'' \emph{Electronics}, vol.~12, no.~17, p. 3676, 2023.

\bibitem{hinton2015distilling}
G.~Hinton, ``Distilling the knowledge in a neural network,'' \emph{arXiv preprint arXiv:1503.02531}, 2015.

\bibitem{Sanh2019DistilBERTAD}
V.~Sanh, L.~Debut, J.~Chaumond, and T.~Wolf, ``Distilbert, a distilled version of bert: smaller, faster, cheaper and lighter,'' \emph{ArXiv}, vol. abs/1910.01108, 2019.

\bibitem{enwiki:1286335380}
\BIBentryALTinterwordspacing
{Wikipedia contributors}, ``Regression analysis --- {Wikipedia}{,} the free encyclopedia,'' 2025, [Online; accessed 22-April-2025]. [Online]. Available: \url{https://en.wikipedia.org/w/index.php?title=Regression_analysis&oldid=1286335380}
\BIBentrySTDinterwordspacing

\bibitem{tf-idf-img}
\BIBentryALTinterwordspacing
C.~Goyal, ``Step by step guide to master nlp – word embedding and text vectorization,'' 2025, [Online; accessed 22-April-2025]. [Online]. Available: \url{https://www.analyticsvidhya.com/blog/2021/06/part-5-step-by-step-guide-to-master-nlp-text-vectorization-approaches}
\BIBentrySTDinterwordspacing

\bibitem{LogisticRegression-scikit}
\BIBentryALTinterwordspacing
SciKit-learn, ``Logisticregression api guide,'' [Online; accessed 22-April-2025]. [Online]. Available: \url{https://scikit-learn.org/stable/modules/generated/sklearn.linear_model.LogisticRegression.html}
\BIBentrySTDinterwordspacing

\bibitem{LogisticRegression-img}
\BIBentryALTinterwordspacing
Niraj, ``Understanding logistic regression from scratch,'' [Online; accessed 23-April-2025]. [Online]. Available: \url{https://nirajpoudel.medium.com/understanding-logistic-regression-from-scratch-82c3ff71f68}
\BIBentrySTDinterwordspacing

\bibitem{mediawiki-api}
\BIBentryALTinterwordspacing
MediaWiki, ``Mediawiki api documentation,'' [Online; accessed 24-April-2025]. [Online]. Available: \url{https://www.mediawiki.org/wiki/API}
\BIBentrySTDinterwordspacing

\bibitem{siino2024text}
M.~Siino, I.~Tinnirello, and M.~La~Cascia, ``The text classification pipeline: Starting shallow going deeper,'' \emph{arXiv preprint arXiv:2501.00174}, 2024.

\bibitem{logistic-regresson-stanford}
D.~Jurafsky and J.~H. Martin, ``Chapter 5: Logistic regression,'' \url{https://web.stanford.edu/~jurafsky/slp3/5.pdf}, 2025, lecture PDF from Stanford University.

\bibitem{hosmer2013applied}
D.~W. Hosmer~Jr, S.~Lemeshow, and R.~X. Sturdivant, \emph{Applied logistic regression}.\hskip 1em plus 0.5em minus 0.4em\relax John Wiley \& Sons, 2013.

\bibitem{enwiki:1287217164}
\BIBentryALTinterwordspacing
{Wikipedia contributors}, ``Generative pre-trained transformer --- {Wikipedia}{,} the free encyclopedia,'' 2025, [Online; accessed 25-April-2025]. [Online]. Available: \url{https://en.wikipedia.org/w/index.php?title=Generative_pre-trained_transformer&oldid=1287217164}
\BIBentrySTDinterwordspacing

\bibitem{islam2024comprehensive}
S.~Islam, H.~Elmekki, A.~Elsebai, J.~Bentahar, N.~Drawel, G.~Rjoub, and W.~Pedrycz, ``A comprehensive survey on applications of transformers for deep learning tasks,'' \emph{Expert Systems with Applications}, vol. 241, p. 122666, 2024.

\bibitem{what-are-transformers}
\BIBentryALTinterwordspacing
A.~AWS, ``What are transformers in artificial intelligence?'' [Online; accessed 27-April-2025]. [Online]. Available: \url{https://aws.amazon.com/what-is/transformers-in-artificial-intelligence/}
\BIBentrySTDinterwordspacing

\bibitem{transformer-attention}
\BIBentryALTinterwordspacing
S.~Cristina, ``The transformer attention mechanism,'' [Online; accessed 27-April-2025]. [Online]. Available: \url{https://machinelearningmastery.com/the-transformer-attention-mechanism}
\BIBentrySTDinterwordspacing

\bibitem{DBLP:journals/corr/LuongPM15}
\BIBentryALTinterwordspacing
M.~Luong, H.~Pham, and C.~D. Manning, ``Effective approaches to attention-based neural machine translation,'' \emph{CoRR}, vol. abs/1508.04025, 2015. [Online]. Available: \url{http://arxiv.org/abs/1508.04025}
\BIBentrySTDinterwordspacing

\bibitem{sklearn}
\BIBentryALTinterwordspacing
{scikit-learn developers (BSD License)}, ``Scikit learn user guide,'' [Online; accessed 09-June-2025]. [Online]. Available: \url{https://scikit-learn.org/stable/user_guide.html}
\BIBentrySTDinterwordspacing

\bibitem{huguet-cabot-navigli-2021-rebel-relation}
\BIBentryALTinterwordspacing
P.-L. Huguet~Cabot and R.~Navigli, ``{REBEL}: Relation extraction by end-to-end language generation,'' in \emph{Findings of the Association for Computational Linguistics: EMNLP 2021}.\hskip 1em plus 0.5em minus 0.4em\relax Punta Cana, Dominican Republic: ACL, Nov. 2021, pp. 2370--2381. [Online]. Available: \url{https://aclanthology.org/2021.findings-emnlp.204}
\BIBentrySTDinterwordspacing

\bibitem{all-mini-lm}
\BIBentryALTinterwordspacing
{Nils Reimers, Omar Espejel, Pedro Cuenca, Tom Aarsen, Arthur Bresnu}, ``all-minilm-l6-v2 sentence-transformer model documentation,'' [Online; accessed 14-June-2025]. [Online]. Available: \url{https://huggingface.co/sentence-transformers/all-MiniLM-L6-v2}
\BIBentrySTDinterwordspacing

\end{thebibliography}

\end{document}